\documentclass[letterpaper]{article} 
\usepackage{aaai24}  
\usepackage{times}  
\usepackage{helvet}  
\usepackage{courier}  
\usepackage[hyphens]{url}  
\usepackage{graphicx} 

\usepackage{amsmath}
\usepackage{amssymb}
\usepackage{color}
\usepackage{booktabs}
\usepackage{multirow}

\usepackage{bbding}

\usepackage{natbib}  
\usepackage{caption} 
\usepackage{algorithm}
\usepackage{algorithmic}

\usepackage{newfloat}
\usepackage{listings}
\DeclareCaptionStyle{ruled}{labelfont=normalfont,labelsep=colon,strut=off} 
\floatstyle{ruled}
\newfloat{listing}{tb}{lst}{}
\floatname{listing}{Listing}
\title{Bi-directional Visual Adapter for Multi-modal Tracking}
\title{Uni-Adapter: }
\title{Parameter-Efficient Mutual Prompt Multi-modal Tracking}
\title{MVP: Mutual Visual Prompt Multi-modal Tracking via Bi-directional Adapter}
\title{Bi-directional Adapter for Multi-modal Tracking}
\author{
    Bing Cao,
    Junliang Guo,
    Pengfei Zhu\thanks{Corresponding author},
    Qinghua Hu
}
\affiliations{

    Tianjin Key Lab of Machine Learning, College of Intelligence and Computing, Tianjin University, China\\
    \{caobing,guojunliang,zhupengfei,huqinghua\}@tju.edu.cn
}

\usepackage{bibentry}

\begin{document}

\maketitle

\begin{abstract}
Due to the rapid development of computer vision, single-modal (RGB) object tracking has made significant progress in recent years. Considering the limitation of single imaging sensor, multi-modal images (RGB, Infrared, etc.) are introduced to compensate for this deficiency for all-weather object tracking in complex environments. However, as acquiring sufficient multi-modal tracking data is hard while the dominant modality changes with the open environment, most existing techniques fail to extract multi-modal complementary information dynamically, yielding unsatisfactory tracking performance. To handle this problem, we propose a novel multi-modal visual prompt tracking model based on a universal bi-directional adapter, cross-prompting multiple modalities mutually. Our model consists of a universal bi-directional adapter and multiple modality-specific transformer encoder branches with sharing parameters. The encoders extract features of each modality separately by using a frozen pre-trained foundation model. We develop a simple but effective light feature adapter to transfer modality-specific information from one modality to another, performing visual feature prompt fusion in an adaptive manner. With adding fewer (0.32M) trainable parameters, our model achieves superior tracking performance in comparison with both the full fine-tuning methods and the prompt learning-based methods. Our code is available: https://github.com/SparkTempest/BAT.

\end{abstract}

\section{Introduction}

Object tracking, a foundation visual task of computer vision, has achieved significant progress over the past decades. Many excellent approaches~\cite{JMMC,ProTrack,ADRNet,DMCNet,zhu2023visual}, and benchmarks~\cite{GTOT,rgbt234,LasHeR,HMFT} have emerged and achieved promising performance on RGB-based object tracking. 
However, due to the imaging mechanism of visible light, some complex scenarios in open environments, such as illumination variation, limit the practical effectiveness of solely RGB-based object tracking, leading to target missing or error tracking.
Different from RGB cameras that capture the reflected light of objects, thermal infrared (TIR) imaging sensors capture the heat emitted by the object itself. 
Compared to RGB images that contain rich color texture in light condition while failing in dark condition,  TIR images provide significant contrast for heat objects while presenting low resolution and poor texture.
Consequently, to conquer the inherent shortcomings of single-modality-based methods, multi-modal object tracking emerged, which fully leveraged RGB and thermal images to perform more robust all-weather tracking.

\begin{figure}[t]
\centering
\includegraphics[width=1\columnwidth]{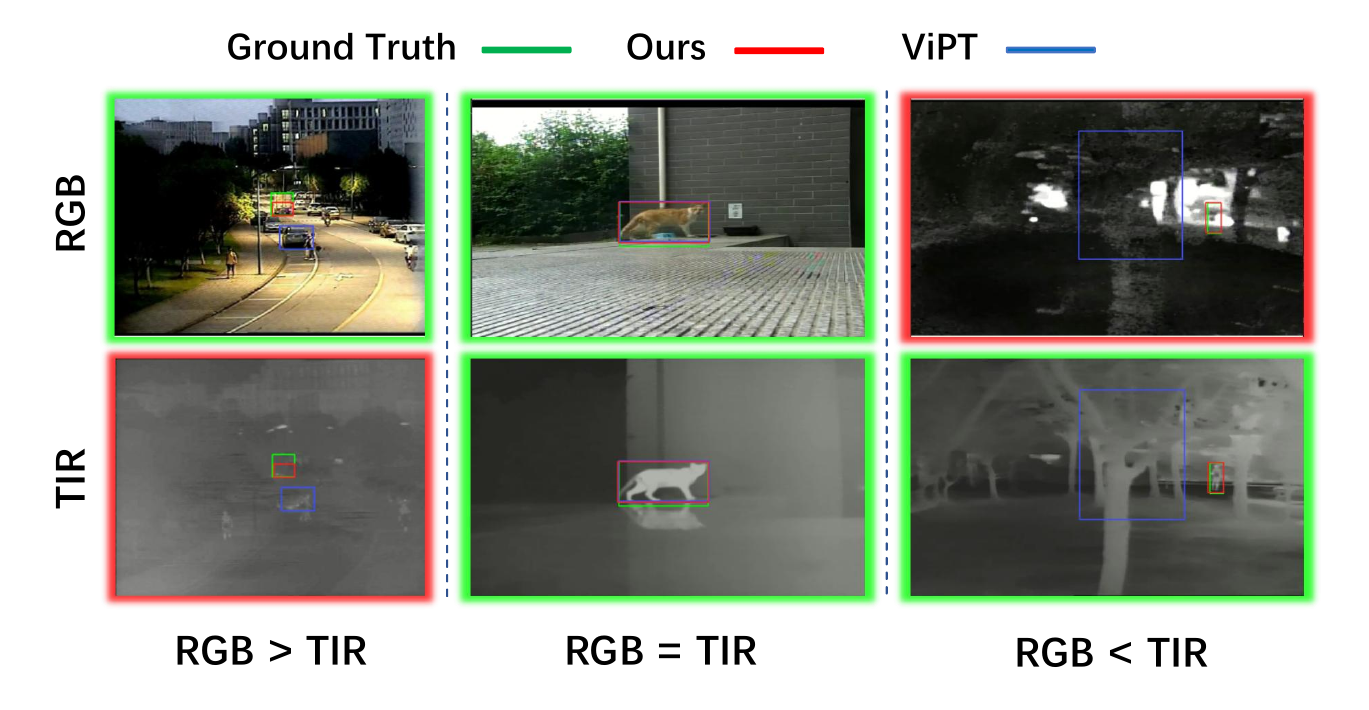} 
\caption{Different dominant modality in complex scenarios. The image with green box represents the dominant modality, and the red box represents the auxiliary modality.
}
\label{fig:dynamic}
\end{figure}

However, existing multi-modal tracking tasks also meet two main issues: i)
Due to the high data labeling cost of multi-modal object tracking, most existing datasets are scale-limited, which is insufficient to support building an effective multi-modal tracker;
ii) The dominant correlation among multi-modal data is not fixed as shown in Fig.~\ref{fig:dynamic}, because different imaging modalities have varying sensitivities to objects in changing environments.

Since pure RGB sequences are much easier to acquire than RGB-T sequence pairs, some multi-modal tracking works~\cite{DAPNet,DAFNet}, accounting for the first limitation, are pre-trained on RGB sequences first and then transferred to multi-modal scenarios in a full fine-tuning manner. For example, The mfDiMP~\cite{mfDimp} takes pre-trained DiMP as foundation models, and fine-tunes it on the generated RGB-T images. Some researchers develop attribute-based multi-modal fusion model~\cite{CAT,ADRNet,APFNet} to reduce reliance on large-scale training data while improving fusion capabilities with a small number of parameters. Despite these methods achieving considerable progress, they also suffered from time expensive and inefficiencies, while showing limited performance. 
In addition to full fine-tuning approaches, 
some recent methods~\cite{ProTrack,zhu2023visual} introduced the parameter-efficient prompt tuning to multi-modal tracking by freezing the backbone parameters and attaching a set of learnable parameters. These methods commonly took one modality (usually RGB) as the dominant modality and another one as the auxiliary modality. However, these methods ignore the dynamic dominant correlation of multi-modal data, making it difficult to fully exploit the complementary multi-modal information in complex scenarios as shown in Fig.~\ref{fig:dynamic}, thus limiting the tracking performance.

In this point, we proposed a Bi-directional Adapter for Multi-modal Tracking (BAT). Different from the methods that add auxiliary modalities as prompts to the dominant modality to enhance the representation ability of the foundation model in downstream tasks (which often use RGB as the primary modality), we do not preset the fixed dominant modality-auxiliary modality, instead BAT dynamically extracts effective information from changing auxiliary modality to dominant modality.
BAT consists of two modality-specific branches and a universal bi-directional adapter. Each modality-specific branch is initialized by the foundation model with fixed parameters during training.
Each modal branch learns the prompt information from the other modality to integrate with the feature information of the current modality, enhancing the representation ability. 
The two modality-specific branch performs interaction by the universal bi-directional adapter to dynamically fuse dominant-auxiliary information mutually in a multi-modal non-fixed association paradigm.
The universal bi-directional adapter has a lightweight hourglass structure. It can be embedded in each transformer layer of the foundation model without introducing a large number of learnable parameters. Experiments on RGBT234~\cite{rgbt234} and LasHeR~\cite{LasHeR} datasets validate the effectiveness of our BAT framework. By training only a few parameters, BAT achieves significant advantages compared with the competing methods.

Our main contributions are summarized as follows:

\begin{itemize}
\item We first propose an adapter-based visual prompt framework for multi-modal tracking. Our model perceives the dynamic changes of the dominant modality in open scenarios, effectively fusing multi-modal information in an adaptive manner.

\item To the best of our knowledge, we for the first time propose a universal bi-directional adapter for the foundation model. It effectively cross-prompts multi-modal tracking with a simple and efficient structure. By only adding 0.32M learnable parameters, our model copes with robust multi-modal tracking in open scenarios.

\item We delved into the effects of our universal adapter on different depths of layers with in-depth analysis.  We also explore even more efficient adapter architecture in experiments, and validated our superiority on multiple RGBT tracking-related datasets against the state-of-the-arts.

\end{itemize}

\section{Related Works}

\subsection{Multi-modal Tracking}
Object tracking is designed to track the assigned initial object in the initial frame and predict its position and scale in subsequent frames. Although numerous excellent studies~\cite{ye2022joint,cui2022mixformer,lan2023procontext} have been proposed and achieved impressive tracking performance, single-modal object tracking is not adequate to meet certain situations, such as low illumination, occlusion, or thermal crossover. Accounting for this, multi-modal tracking has gained increased attention because different modalities have the potential to offer complementary information mutually, boosting tracking performance in challenging scenarios that are difficult to handle purely by single-modal images. For example, FANet~\cite{FANet} design a feature aggregation module to fuse multi-modal features within each modality and an adaptive aggregation module to fuse multi-modal features in different modalities. HMFT~\cite{HMFT} design a hierarchical fusion framework to integrate multi-modal features. APFNet~\cite{APFNet} used an attribute-based fusion framework to aggregate attribute-specific fusion features, with a transformer structure to strengthen the multi-modality features.

\begin{figure*}[ht]
\centering
\includegraphics[width=1\textwidth]{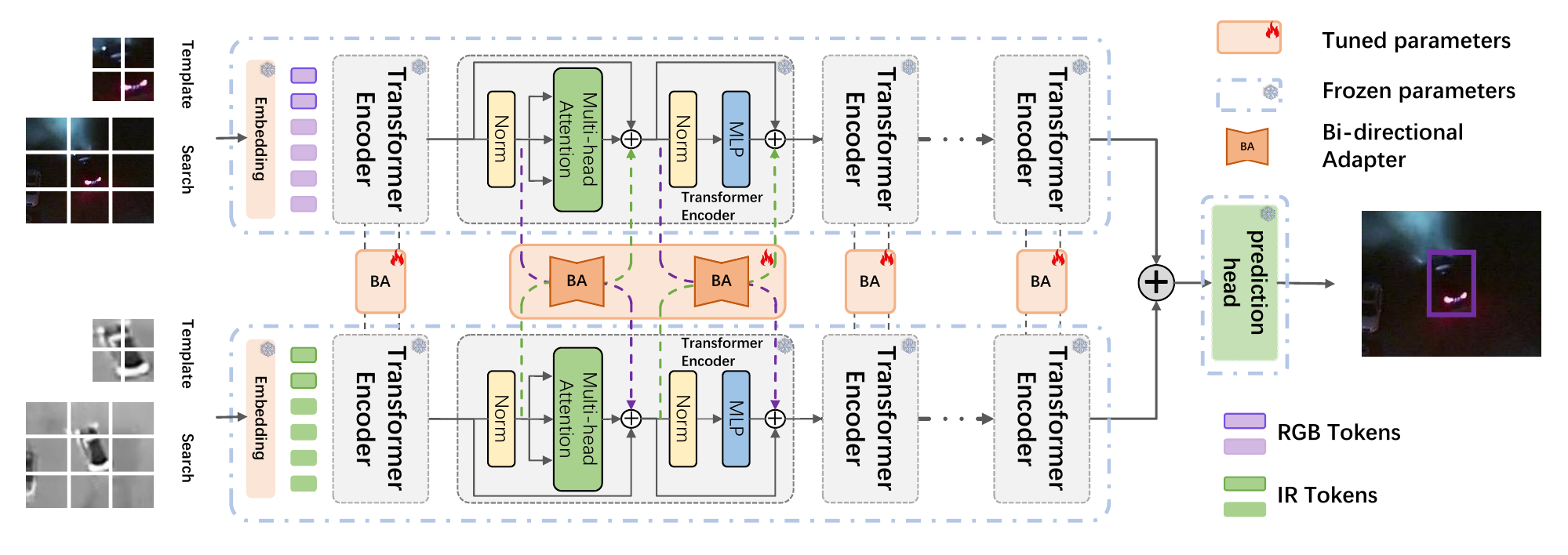} 
\caption{The overall architecture of our proposed BAT. We first transformed the template frame and search frame of each modality into tokens, then concatenated them together to pass the $N$-layer dual-stream transformer encoder, respectively. 
The bi-directional adapter is paralleled with the dual-stream encoder layer, which could learn feature prompts from one modality to another. To this end, the output features of the two branches are added and fed into the prediction head for final tracking result.
}
\label{fig:framework}
\end{figure*}

\subsection{Parameter-Efficient Tuning}
Fine-tuning is a widely studied technique over the past decades that usually transfers large pre-trained models to downstream tasks by updating all the parameters on task-oriented data.
These full fine-tuning method are parameter-inefficient and also require sufficient data to optimize all the parameters.
Recently, prefix-tuning, a new paradigm for parameter-efficient tuning, has become widely employed in natural language processing (NLP), which has demonstrated its efficiency in a variety of extended computer vision tasks~\cite{khattak2023maple}.
VPT~\cite{jia2022visual} introduces prompt-tuning into the vision task, adding learnable tokens from the input layer and freezing the backbone to train the classification head and the newly added prompt token, achieving better results than full-tuning-based methods. 
Protrack~\cite{ProTrack} provides a new perspective for multi-modal tracking by transforming multi-modal inputs into a single modality through a prompt paradigm. It exploits the tracking ability of pre-trained RGB trackers, rather than building complex multi-modal fusion modules.
Inspired by this, ViPT~\cite{zhu2023visual} designs a learnable prompts generation module to generate prompts for RGB modality based on thermal infrared modality on downstream tasks.
Unlike visual-language models, RGB-T tracking employs two comparable modalities, which can be both contributed by the pre-trained visual foundation model. However, previous methods mainly take RGB as the dominant modality, ignoring the dynamically changing environments where TIR has stronger representation ability than RGB, performing the dominant modality. 
This motivates us to break away from the fixed multi-modal correlation paradigm and design a universal bi-directional adapter that does not predefine the dominant modality and can adaptively extract features from RGB and IR.

\section{Methodology}
In this paper, we propose a novel universal bi-directional adapter for multi-modal tracking (BAT), which cross-prompts multi-modal data mutually. 
Instead of fully fine-tuning the foundation model, BAT transfers the pre-trained tracker to multi-modal scenarios effectively and efficiently by only learning the lightweight adapter, performing excellent multi-modal complementarity and superior tracking accuracy.
We present the overall architecture of our BAT in Fig.~\ref{fig:framework}.

\subsection{Multi-modal Tracking}
Given a video $\mathcal{V}$ with an initial box position $\mathcal{B}_0$ of the target object $\mathcal{Z}_0$ in the first frame $\mathcal{I}_{template}$, single-model object tracking learns to search for this object in the subsequent frames $\mathcal{I}_{search}$.
Typically, the object tracker $\mathcal{T}$ consists of a feature extraction function $\mathcal{F}$ and a head box $\mathcal{H}$. 
For the transformer-based foundation model, the template frame $\mathcal{I}_{template}$ and the search frame $\mathcal{I}_{search}$ are transformed into tokens by patch embedding and position embedding, and then concatenated together to pass through $N$-layer transformer encoder for joint feature extraction. 
Finally, the output token of encoder for the corresponding search image is fed into the prediction head to obtain the target tracking result. Thus, the position of the box $\mathcal{B}$ in subsequent frames is predicted by
\begin{equation}
    \mathcal{B}=\mathcal{H}(\mathcal{F} (\mathcal{I}_{template}, \mathcal{I}_{search}, \mathcal{B}_0)),
\end{equation}
where $\mathcal{F}$ is a pre-trained transformer backbone with powerful representation ability.

Multi-modal tracking (MMT) extends this setting to multiple videos in different modalities by formally introducing another modal stream, which jointly makes the final decision for the tracking objects.
Take RGB-T as an example, the RGB and thermal modalities are temporally synchronized and spatially aligned. MMT tracks $\mathcal{Z}_0$ from both the subsequent frames of both RGB modality $\mathcal{I}_{search}^{RGB}$ and TIR modality $\mathcal{I}_{search}^{T}$ as,
\begin{equation}
    \mathcal{B}=\mathcal{H}(\mathcal{F} (\mathcal{I}_{template}^{RGB}, \mathcal{I}_{template}^{T}, \mathcal{I}_{search}^{RGB}, \mathcal{I}_{search}^{T}, \mathcal{B}_0)).
\end{equation}

As shown in Fig.~\ref{fig:framework}, our BAT has a dual-stream encoder structure for RGB modality and thermal infrared modality respectively, each stream of which shares the same parameters. BAT first feeds the two modalities $\mathcal{I}_{template}^{RGB}$, $\mathcal{I}_{search}^{RGB}$ and $\mathcal{I}_{template}^{T}$, $\mathcal{I}_{search}^{T}$ to a patch and position embedding layer, and obtain the RGB tokens $x^{RGB}_0$ and TIR tokens $x^{TIR}_0$. Then, our universal bi-directional adapter is embedded in the $i$-th layer of transformer encoder, penetrating the encoders of two modalities. For the $i+1$ layer of each encoder, it learns to integrate the modality-specific feature with complementary information of another modality from its previous layer.  Each encoder learns feature prompts from another modality in a layer-by-layer manner,
\begin{equation}
    (x_{i+1}^{RGB},x_{i+1}^{TIR}) = \mathcal{F}_i^{A}( x_{i}^{RGB}, x_{i}^{TIR}),              i = 1, 2, \cdots, N,
\end{equation}
where $\mathcal{F}_i^{A}$ refers to the dual-stream encoder layer paralleled with our bi-directional adapter structure.

To this end, the multi-modal feature of tracking objects is progressively and dynamically extracted during the $N$ layers of the transformer encoder in the foundation model.
Finally, the features of two modal branches are added and fed into the prediction head to obtain the final tracking results as
\begin{equation}
    \mathcal{B}_{box} = \mathcal{H}(x_{N}^{RGB} + x_{N}^{TIR}), i = 1, 2, \cdots, N.
\end{equation}

\begin{figure}[ht]
\centering
\includegraphics[width=1\columnwidth]{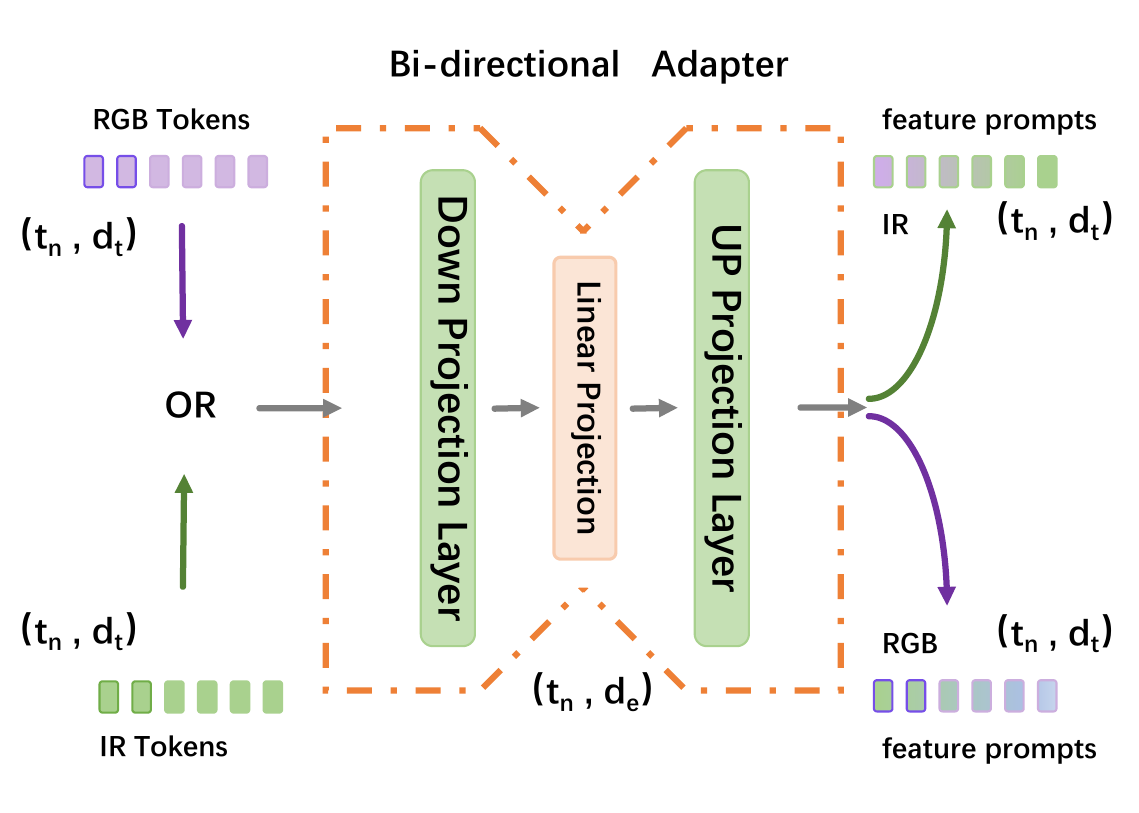}
\caption{The detailed architecture of bi-directional adapter. 
It consists of three linear projection layers, $t_{n}$ represents the token's num of each modality, the input token is first dimensional reduced to $d_{e}$ and passed through a linear projection layer, then up-project to the original dimension $d_{t}$ and fed into another modality as feature prompts.
}
\label{fig:adapter}
\end{figure}

\subsection{Bi-directional Adapter}

Our bi-directional adapter (BAT) is designed to cope with transferring complementary features from one modality to another in a universal manner. 
The input modality is self-adaptive, capable of dynamically extracting the features of the auxiliary modality and transferring them to the dominant modality as the environment changes.

As shown in Fig.~\ref{fig:framework}, the bi-directional adapter adopts a modular design, which is embedded in the multi-head self attention stage and the MLP stage, separately. Here, we take the processing of $x_{i}^{RGB}\rightarrow x_{i}^{TIR}$ as an example to detail our bi-directional adapter. The $i$-th layer of RGB branch integrates the auxiliary information from TIR branch through the adapter as,

\begin{equation}
\begin{aligned}
   {x_{i}^{RGB}}' = x_{i}^{RGB} + \mathcal{F}^{Att}(x_{i}^{RGB}) + P_{i}^{TIR},\\
    P_{i}^{TIR} = \mathcal{F}^{Ada}(x_{i}^{TIR}),   i = 1,2,...,N 
\end{aligned}
\end{equation}
where $\mathcal{F}^{Att}$ and $\mathcal{F}^{Ada}$ represent the multi-head self-attention block and our bi-directional adapter network, respectively. $F^{Ada}(\cdot)$ refers to the output feature prompt of the bi-directional adapter.
The $P_{i}^{TIR}$ is the feature prompt extracted from TIR modality.  $x_{i}^{RGB}$ is fed to the multi-head self attention block after passing a layernorm operator, and then added with $x_{i}^{RGB}$ and $P_{i}^{TIR}$ to obtain ${x_{i}^{RGB}}'$. 

In the next stage, ${x_{i}^{RGB}}'$ is fed into the multi-layer perceptron $F^{MLP}$, and added with feature prompt ${P_{i}^{TIR}}'$ and ${x_{i}^{RGB}}'$ together to obtain the output $x_{i+1}^{RGB}$ of the $i+1$ layer in the RGB encoder.

\begin{equation}
     x_{i+1}^{RGB} = {x_{i}^{RGB}}' + F^{MLP}({x_{i}^{RGB}}') + {P_{i}^{TIR}}', 
\end{equation}

\begin{equation}
    {P_{i}^{TIR}}' = F^{Ada}({x_{i}^{TIR}}'),   i = 1,2,...,N.
\end{equation}

The detailed architecture of our bi-directional adapter is depicted in Fig.~\ref{fig:adapter}, which is designed to transfer feature prompts from one modality to another modality. 
The input token of bi-directional adapter block is first reduced to $d_e$ dimension by down projection layer, and then passed through a linear projection layer.
Then, it is up-projected to the original dimension, and fed back to transformer encoder layer of the other modality as the feature prompt. Through this simple structure, bi-directional adapter effectively perform feature prompting between $x_{i}^{RGB}$ and $x_{i}^{TIR}$ modalities for multi-modal tracking.

As for freezing the transformer encoder and prediction head, we only need to optimize a few parameters of the newly added adapter. It is worth noting that, different from most conventional adapters, our bi-directional adapter is performed as a cross-modal feature prompt for the dynamically changing dominant modality, ensuring promising tracking performance in the open world.

\subsection{Objective Loss} 

The token sequence is first converted to a 2D spatial feature map, using a series of fully convolutional networks (FCN), and outputs the target classification score map (indicating the target location), offset, and the normalized bounding box.
The overall loss function of BAT is formulated as,

\begin{equation}
L_{total} = L_{cls} + \lambda_1L_{iou} + \lambda_2L1.
\end{equation}
where $L_{cls}$ denotes the weighted focal loss for classification, the generalized IoU loss $L_{iou}$ and $L_1$ are adopted for bounding box regression, ${\lambda}_{1}$ and ${\lambda}_2$ are trade-off parameters.

\section{Experiments}

\subsection{Experimental Setting}
\subsubsection{Datasets and Evaluation Metrics.}
We conduct experiments on two multi-modal tracking datasets: RGBT234~\cite{rgbt234} and LasHeR~\cite{LasHeR}, and evaluate the tracking performance with four evaluation metrics: Precision Rate (PR), Maximum Precision Rate (MPR), Success Rate (SR), and Maximum Success Rate (MSR).

\textbf{RGBT234} provides 234 sequences of aligned RGB and infrared videos. It offers 12 attributes, including LI (Low Illumination), Occlusion, DEF (Deformation), Movement, etc. The total number of frames is about 234K, with a maximum of 8K frames per sequence.  It provides the ground-truth label for both RGB and TIR modalities, allowing trackers to perform multi-modal performance evaluations.

Due to the RGBT234 use of a parallel optical axis visible light-infrared imaging system, no pre-processing or post-processing (such as stereo matching and color correction) is required. Its cross-modal alignment is more accurate, but the ground truth for RGB and IR is still not completely consistent. Therefore, for a fair comparison, we use the MPR and MSR instead of PR and SR as the evaluation metrics. Specifically, for each frame, the Euclidean distance between the result box and the ground truth is calculated separately in the RGB and IR modalities, and the smaller distance is used to calculate the accuracy.

\textbf{LasHeR} is an RGBT tracking dataset that contains 1224 RGBT sequences with 
730K frames, captured in various types of imaging platforms. It includes 19 video attributes, adding 7 types of new attributes such as ``HI'' (High Illumination), ``FL'' (Frame Lost), and ``AIV'' (Abrupt Illumination Variation) on the basis of previous ones, making it an even more challenging dataset for RGBT tracking tasks.

To address the alignment issue of radial distortion images in different RGBT modalities, the LasHeR dataset only performs precise alignment on the local area covering the target object in each frame, since the object tracking task does not emphasize the tracking effect of the background. Thus, for each frame, a set of matching points is labeled to transform its RGB image into the same coordinate system as the thermal infrared image. This results in consistent ground truth for both modalities. Different from the RGBT234 dataset, PR and SR can be used as evaluation metrics.

\subsubsection{Implementation Details.}
We implement our BAT based on the Pytorch and train it on 4 NVIDIA RTX A6000 GPUs with a batch size of 32.  
We follow the hyper-parameters setting of the foundation model in the loss function. The AdamW optimizer~\cite{loshchilov2019adamw} with a weight decay of $10^{-4}$ is adopted, and the learning rate is set to $4\times 10^{-4}$.
The fixed parameters of the modal-specific branch in BAT are initialized by the pre-trained foundation model~\cite{ye2022joint}. The fine-tuning of our BAT on the LasHeR training set takes 60 epochs for 8 hours, where each epoch contains $6\times 10^{4}$ sample pairs.

\subsection{Comparisons}
We compare our model with 19 competing methods. The quantitative comparisons are reported in Table~\ref{tab:comparison} and the qualitative evaluation results are presented in Fig.~\ref{fig:comparison}.

\subsubsection{Quantitative Evaluation on RGBT234.}
As shown in Table~\ref{tab:comparison}, for full-tuning competing methods, DMCNet~\cite{DMCNet} achieved considerable performance with the runner-up MPR score of 83.9\%. While the SOTA efficient-tuning methods ViPT~\cite{zhu2023visual} achieved similar performance as DMCNet, with a worse MPR score of 83.5\% and a slightly higher MSR score of 61.7\%. The existing efficient-tuning methods did not perform significant improvements. This may be due to that these efficient-tuning methods are challenging to dynamically extract compatible information from both RGB and TIR modalities. As a comparison, our BAT achieves 86.8\% MPR and 64.1\% MSR, outperforming the runner-up MPR and MSR scores by 2.9\% and 2.4\% respectively, which is a significant improvement among all the competing methods. The experimental results demonstrate the effectiveness of our BAT model.

\begin{table}[th]
\label{tabresults}
\renewcommand{\arraystretch}{1}
\setlength{\tabcolsep}{6pt}
\caption{Overall performance on RGBT234 and LasHeR dataset. \textcolor{red}{Red}/\textcolor{green}{Green}/\textcolor{blue}{Blue} indicates the best/runner-up/third best results. Results are reported in percentage (\%).}
\label{tab:comparison}
\centering
{
\begin{tabular}{lcccc}
\toprule
&\multicolumn{2}{c}{RGBT234} &\multicolumn{2}{c}{LasHeR} \\
Method&MPR$\uparrow$&MSR $\uparrow$&PR $\uparrow$&SR $\uparrow$ \\
\midrule
ATOM~(\citeyear{atom})&-&-&40.6&30.7 \\
DiMP-50~(\citeyear{dimp50})&-&-&44.2&33.6 \\
mfDiMP~(\citeyear{mfDimp})&64.6&42.8&44.8&34.3 \\
DAPNet~(\citeyear{DAPNet})&76.6&53.7&43.1&31.4 \\
SiamFC++~(\citeyear{siamfc++})&-&-&34.8&27.4 \\
CAT~(\citeyear{CAT})&80.4&56.1&45.0&31.4 \\
CMPP~(\citeyear{CMPP})&82.3&57.5&-&- \\
STARK ST-50~(\citeyear{stark})&-&-&44.9&36.1 \\
TransT~(\citeyear{transt})&-&-&52.4&39.4  \\
JMMAC~(\citeyear{JMMC})&79.0&57.3&-&- \\
MANet++~(\citeyear{MANet++})&79.5&55.9&46.7&31.4 \\
FANet~(\citeyear{FANet})&78.7&55.3&44.1&30.9 \\
ADRNet~(\citeyear{ADRNet})&80.9&57.1&-&- \\
OSTrack-256~(\citeyear{ye2022joint})&72.9&54.9&51.5&41.2 \\
APFNet~(\citeyear{APFNet})&82.7&57.9&50.0&36.2 \\
DMCNet~(\citeyear{DMCNet})&\textbf{\textcolor{green}{83.9}}&59.3&49.0&35.5 \\
HMFT~(\citeyear{HMFT})&78.8&56.8&-&- \\
\midrule
ProTrack~(\citeyear{ProTrack})&79.5&\textbf{\textcolor{blue}{59.9}}&\textbf{\textcolor{blue}{53.8}}&\textbf{\textcolor{blue}{42.0}} \\
ViPT~(\citeyear{zhu2023visual})&\textbf{\textcolor{blue}{83.5}}&\textbf{\textcolor{green}{61.7}}&\textbf{\textcolor{green}{65.1}}&\textbf{\textcolor{green}{52.5}} \\
\midrule
\textbf{BAT (Ours)}&\textbf{\textcolor{red}{86.8}}&\textbf{\textcolor{red}{64.1}}&\textbf{\textcolor{red}{70.2}}&\textbf{\textcolor{red}{56.3}} \\
\bottomrule
\end{tabular}
}
\end{table}

\subsubsection{Quantitative Evaluation on LasHeR.}
Compared with the RGB234 dataset, the LasHeR dataset is more challenging, due to more extreme attributes being introduced. The performance gap of most existing methods is significantly widened.
Previous advanced methods such as DMCNet and APFNet perform unsatisfactory on this dataset. 
Even OSTrack, a tracker only based on RGB modality, reaches a stronger performance than many RGBT trackers, which is completely opposite to the situation where multi-modal trackers take the lead in the RGBT234 dataset.
The efficient-tuning methods such as ProTrack and ViPT are significantly superior to the traditional methods, which may benefit from the strong representation ability of the used pre-trained foundation model. 
As shown in Table~\ref{tab:comparison}, ViPT achieved 65.1\% of PR and 52.5\% of SR, which is a considerable improvement among the competing methods. However, our BAT even further improves the ViPT at 5.1\% and 3.8\%, reaching 70.2\% and 56.3\% of PR and SR scores, respectively. This experiment further validates our universal bi-directional adapter in learning dynamically changing attributes in complex environments.

\begin{figure*}[ht]
\centering
\includegraphics[width=1\textwidth]{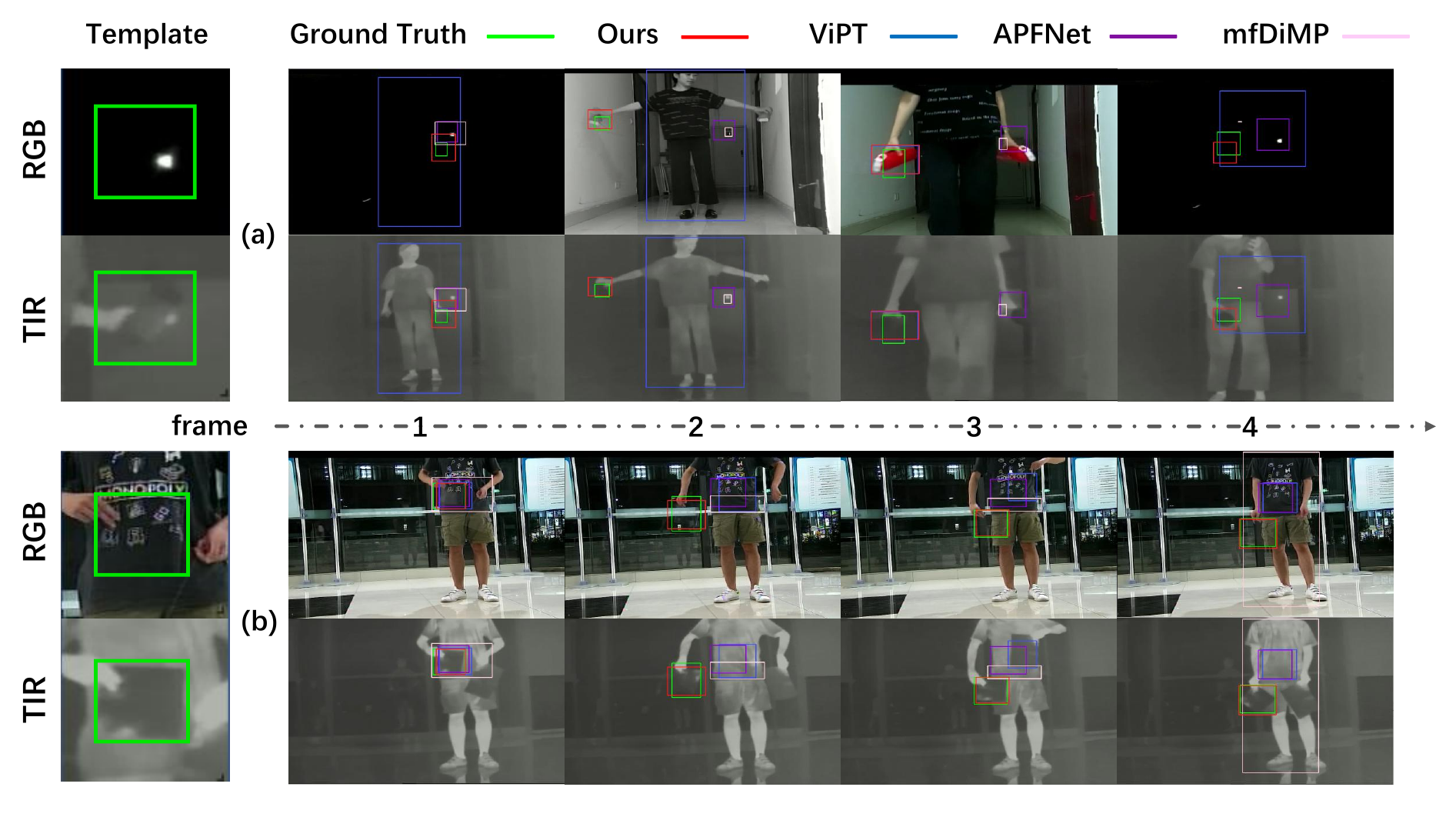}
\caption{Visualization of tracking results. The green rectangles indicate target objects in the template frame. Our method shows the best performance in different frame sequences as the dominant modality dynamically changes.}
\label{fig:comparison}
\end{figure*}

\begin{figure}[ht]
\centering
\includegraphics[width=1\columnwidth]{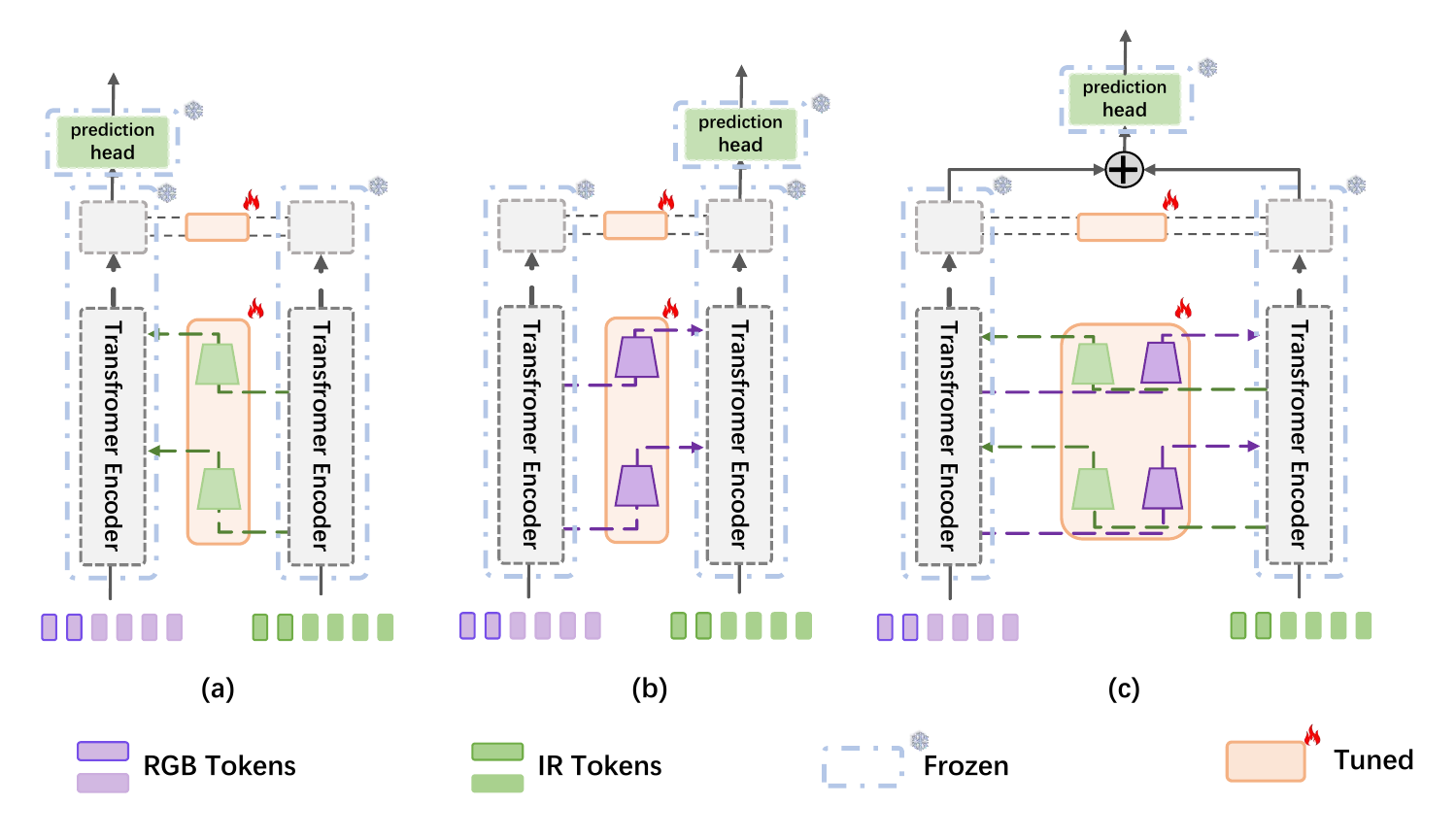}
\caption{Different variants of bi-directional adapter for dual-stream encoder framework. }
\label{fig:adaptervariant}
\end{figure}

\begin{figure}[ht]
\centering
\includegraphics[width=1\columnwidth]{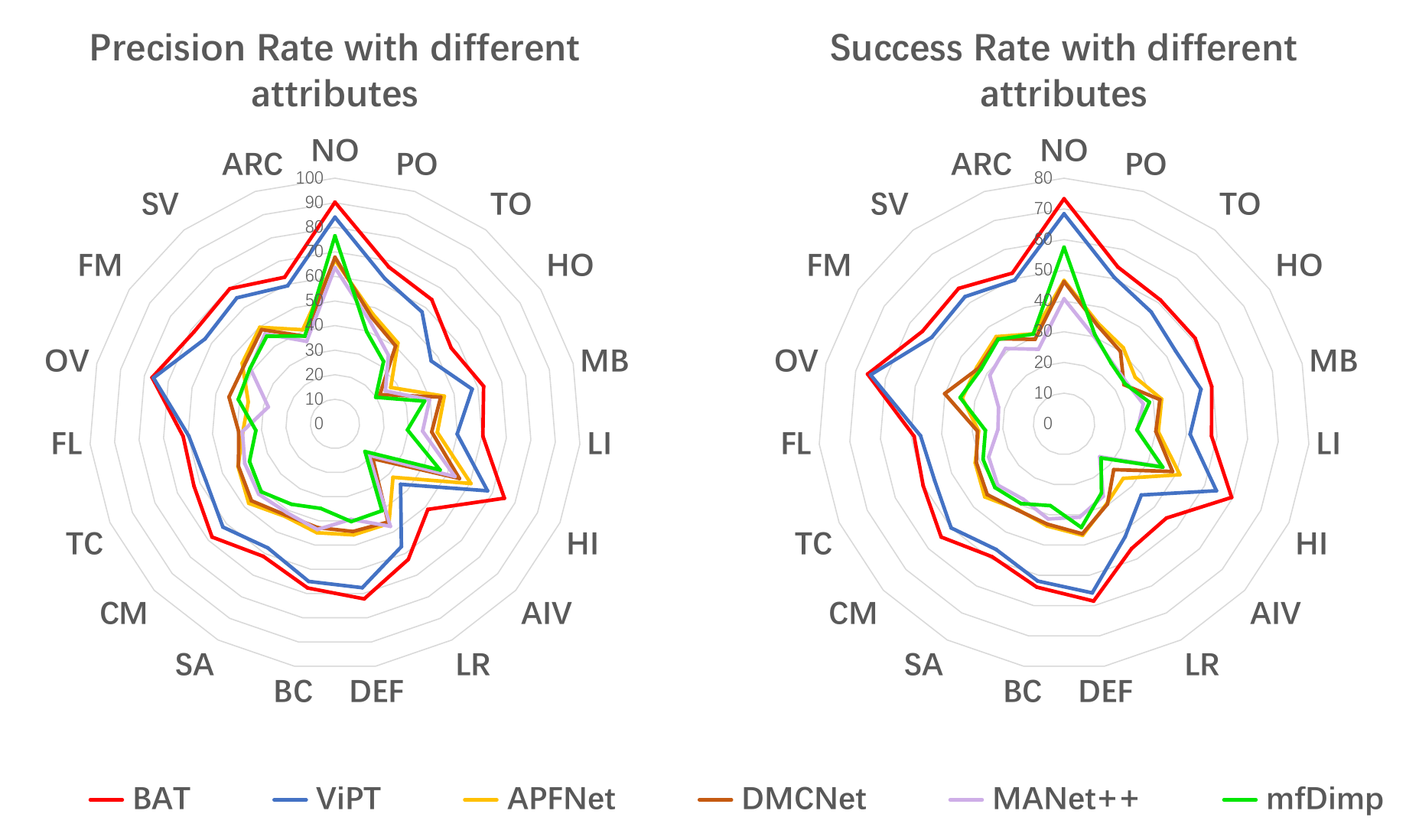} 
\caption{More comparisons of BAT and the competing methods under different attributes in the LasHeR dataset.}
\label{fig:attribute}
\end{figure}

\subsubsection{Qualitative Evaluation.}
Since our dual-stream encoder does not rely on a single modality as the dominant modality, it outperforms single-stream prompt-learning approaches in complex scenarios when RGB images are distorted or even unavailable.
As shown in Fig.~\ref{fig:comparison}(a), the tracking information provided by the video sequence in the early stage strongly depends on the TIR image, and after a few frames, the RGB image progressively dominates and provides more effective information than TIR. Fixed correlation methods, such as ViPT, mainly using RGB as the dominant modality, cope with tracking the object in subsequent bright light scenes while failing to dynamically track the accurate position in dark light environments. Our method effectively tracks the target even when RGB is completely unavailable, and the tracking results are much better when both RGB and TIR provide effective information in subsequent scenes.
As shown in Fig.~\ref{fig:comparison}(b), for transparent object tracking, the features provided by the RGB modality in this scenario have strong interference. Compared with other methods that fail to track, our bi-directional adapter dynamically extracts effective features of the target from both RGB and IR modalities, capturing a more accurate target response position, and eliminating the interference of the RGB modality.
These experiments demonstrate our effectiveness in dynamically prompting effective information from the changing dominant-auxiliary modalities in complex scenarios.

\subsection{Discussion}

\subsubsection{Effect of Different Adapter Variants.}

In this section, we explore the different adapter variants in our BAT framework. As shown in Fig.~\ref{fig:adaptervariant}, the adapter can be performed in two single directions: $RGB\rightarrow TIR$ in Fig.~\ref{fig:adaptervariant}(a) and $TIR\rightarrow RGB$ in Fig.~\ref{fig:adaptervariant}(b).
The single-directional adapter only extracts feature prompts from one modality to another, and only takes one stream's transformer encoder layer's output to regression final result.
Fig.~\ref{fig:adaptervariant}(c) presents the dual-adapter architecture without sharing parameters. Each adapter only extracts feature prompts from one specific modality to another modality.
We use the foundation model as our baseline. 
The dual-stream framework initialized by the parameters of foundation model is denoted as Baseline-Dual,
which takes the specific stream that has the maximum score-map value in the prediction head to calculate the final result.

We reported the results of different variants of adapter in Table~\ref{tab:adaptertype}. 
The dual-stream baseline (Baseline-Dual) is slightly better than the foundation model (Baseline), which demonstrates the foundation model has the potential to be applied in both RGB and TIR modalities.
For the single-directional adapters, BAT-RGB and BAT-TIR achieve a significant improvement over the baseline models. This might be due to that the effective information of one modality is transferred to the other modality through the adapter. It further validated that multi-modal data provides more complementary information than single modality. Meanwhile, the difference between BAT-RGB and BAT-TIR is small (less than 3\%), indicating that the dominant correlation is not fixed, dynamically learning from the two modalities has the potential to perform even better tracking results in more complex conditions.
The dual-adapter (BAT-Dual) requires double parameters compared to our universal bi-directional adapter, while maintaining the similar performance as our universal version. This may be because BAT adopts the same foundation model for two modality-specific branches that are fixed during training. Therefore, the feature distribution should be compatible with both modalities to dynamically balance the changing dominant and auxiliary modalities.
Our bi-directional adapter cross-prompts the two branches with a universal adapter, learning compatibility of multiple modalities and achieving comparable performance with half learnable parameters. Our universal adapter is also more flexible to handle more modalities in a parameter-efficient manner.

\begin{table}[t]
    \centering
    \setlength{\tabcolsep}{12pt}
    \caption{Quantitative comparison between different variants of BAT on the LasHeR dataset}
    \begin{tabular}{lcc}
        \toprule
        Method        & PR& SR  \\
        \midrule
        Baseline     &   51.5  & 41.2\\
        Baseline-Dual    & 52.2   & 42.8   \\
        BAT-RGB & \textbf{\textcolor{blue}{69.0}}  & \textbf{\textcolor{blue}{55.4}}   \\
        BAT-TIR&   68.5 & 54.8 \\
        BAT-Dual & \textbf{\textcolor{green}{69.6}} & \textbf{\textcolor{red}{56.4}}   \\
        \midrule
        BAT          &   \textbf{\textcolor{red}{70.2}} &\textbf{\textcolor{green}{56.3}} \\
        \bottomrule
    \end{tabular}
    \label{tab:adaptertype}
\end{table}

\subsubsection{Effect of Adapter in Different Layers.}

To explore a more efficient adapter architecture, we choose parts of transformer encoder layers to embed our bi-directional adapter.
We preset 6 embedding types, formulated as BAT-``n'', where ``n'' represents the remaining number of adapter layers.
The results are shown in Table~\ref{tabadapternum}. The layers indicate the position of our bi-directional adapter in the foundation model. 
The performance of BAT-1 is limited as few multi-modal information cross-prompts during training.
BAT-4 in the middle layers achieves comparable performance as BAT-12, while saving more learnable parameters. It demonstrates our universal bi-directional adapter has the potential to be further simplified.

\begin{table}[t]
\caption{Results of different bi-directional adapter layers in LasHeR}
\centering
\begin{tabular}{ccccc}
\toprule
Type&Layers&PR&SR \\
\midrule
BAT-1&1&  61.4  &49.3 \\
BAT-1&12&  61.6 &49.9 \\
BAT-4&1-4&  65.2&52.5 \\
BAT-4&5-8&  \textbf{\textcolor{green}{68.6}}&\textbf{\textcolor{green}{55.2}} \\
BAT-4&9-12&\textbf{\textcolor{blue}{66.4}} &\textbf{\textcolor{blue}{53.4}} \\
BAT-12&1-12&\textbf{\textcolor{red}{70.2}} &\textbf{\textcolor{red}{56.3}} \\
\bottomrule
\end{tabular}
\label{tabadapternum}
\end{table}

\begin{table}[t]
\caption{More quantitative comparisons of PR and SR scores under five extreme attributes in LasHeR dataset}
\label{tab:attribute}
\centering
\begin{tabular}{ccccc}
\toprule
Attribute&mfDiMP&APFNet&ViPT&Ours \\
\midrule
 NO&76.5/57.5&66.7/46.7&84.1/68.4&90.2/73.3 \\
 HO&19.8/23.8&27.1/27.7&46.8/43.4&56.5/51.0 \\
 LI&29.6/23.8&41.8/30.8&49.9/41.2&60.4/48.2  \\
 AIV&16.6/16.4& 32.1/26.2&36.3/34.2&51.4/45.3  \\
 TC&38.0/28.8& 43.1/31.6& 57.4/46.0&62.7/50.1 \\
\bottomrule
\end{tabular}
\end{table}

\subsubsection{More Comparisons under Different Attributes.}
Since the LasHeR dataset provides 19 additional attributes 
in addition to the point annotates, we further evaluate the proposed BAT with some advanced competing methods on each attribute. As shown in Fig~\ref{fig:attribute}, our BAT outperforms the competing methods in all the attributes. This experiment not only validates the effectiveness of our method, but also demonstrates our superior generalization under different conditions. Moreover, we further explore the performance of BAT in five extreme attributes such as ``NO'', ``HO'', ``LI'', ``AIV'' and ``TC''. The experimental results of PR and SR scores are reported in Table~\ref{tab:attribute}. Compared with the comparisons on the complete LasHeR dataset, our model outperforms the competing methods even more under extreme attributes. Our PR score of ``LI'' attribute and SR score in ``AIV'' attribute  surpasses ViPT by over 10\%. These attributes show more dynamic than other attributes, which further verifies the effectiveness of our bi-directional adapter in dynamically learning the changing dominant-auxiliary information for multi-modal tracking in complex environments.

\section{Conclusion}

In this work, we present BAT, a new bi-directional adapter by introducing a universal feature prompt-learning paradigm to multi-modal tracking. 
The core idea of BAT is to dynamically excavate the changing dominant-auxiliary relevance of multiple modalities in complex scenarios, and extract complementary information from the pre-trained foundation model.
Extensive experiments on multiple RGBT tracking datasets demonstrate the superiority of BAT over competing methods. With the in-depth study of the adapter structure of BAT, we believe this work has the potential to be applied to broader tasks. We expect it can attract more attention to multi-modal parameter-efficient tuning and empower, more general, vision-language tasks. Moreover, our method is currently validated in the RGB and TIR tracking task, and in the future, we are interested in exploring a general model for more diverse modalities.



\bibliography{aaai24}

\end{document}